\documentclass[twoside,11pt]{article}
\usepackage{setspace}
\usepackage{obs_study_style}

\usepackage[utf8]{inputenc}
\usepackage{amsmath}
\usepackage{setspace}
\usepackage{amssymb}
\usepackage{comment}
\usepackage{graphicx}
\usepackage{tikz}
\usepackage{soul}
\usepackage{natbib}
\usetikzlibrary{matrix}

\heading{}{}{}{}{}{Chengliang Tang, Gan Yuan, and Tian Zheng}

\ShortHeadings{Weakly Supervised Learning Creates a Fusion of Modeling Cultures}{Tang, Yuan, and Zheng}
\firstpageno{1}

\begin{document}

\title{Weakly Supervised Learning Creates a \\ Fusion of Modeling Cultures}

\author{\name Chengliang Tang \email ct2747@columbia.edu \\
        \name Gan Yuan \email gy2277@columbia.edu \\
        \name Tian Zheng \email tian.zheng@columbia.edu \\
        \addr Department of Statistics\\
       Columbia University\\
       1255 Amsterdam Avenue, New York, NY 10027 
       }

\maketitle

\begin{abstract}
The past two decades have witnessed the great success of the {\em algorithmic modeling} framework advocated by \citet{breiman2001statistical}. Nevertheless, the excellent prediction performance of these black-box models rely heavily on the availability of {\em strong supervision}, i.e. a large set of accurate and exact ground-truth labels. In practice, strong supervision can be unavailable or expensive, which calls for modeling techniques under {\em weak supervision}. In this comment, we summarize the key concepts in weakly supervised learning and discuss some recent developments in the field. Using algorithmic modeling alone under a weak supervision might lead to unstable and misleading results. A promising direction would be integrating the {\em data modeling} culture into such a framework.  
\end{abstract}


As an important think piece to both the statistics and machine learning communities, \citet{breiman2001statistical} laid out the contrast of the two cultures in modeling thinking: {\em data modeling} and {\em algorithmic modeling}. It pointed out the limitations of data modeling and the opportunities and potentials of algorithmic modeling. Over the past two decades, \citet{breiman2001statistical}'s vision for algorithmic modeling has been validated by the rapid development and application of complicated yet effective algorithmic models, e.g., deep learning. Meanwhile, new challenges and opportunities are emerging everyday as we continue to deal with data with increasing size and complexity. In this comment, we offer a brief discussion of recent developments in the field of {\em weakly supervised learning} and discuss how it creates a need for {\em data modeling} thinking in an {\em algorithmic modeling} framework.

Following the taxonomy introduced in \citet{breiman2001statistical}, the {\em data modeling} culture refers to methods that explicitly assume a stochastic model for data generation. Often, methods of this culture have shallow structures and are easy to interpret. Typical examples include linear regression, logistic regression, to name a few. The validity of such methods is backed by the probabilistic properties of their outputs, such as goodness-of-fit tests and residual analyses.  In contrast, the {\em algorithmic modeling} culture aims to learn the complex and unknown nature of true data generation mechanisms through ``black-box'' algorithms. Typical examples of this culture include decision trees, support vector machines (SVM), and neural networks (NN). The training and evaluation of these algorithms are guided by predictive accuracy. 

\begin{figure}[h]
\centering
\vspace{.3cm}
\includegraphics[height=0.18\textheight]{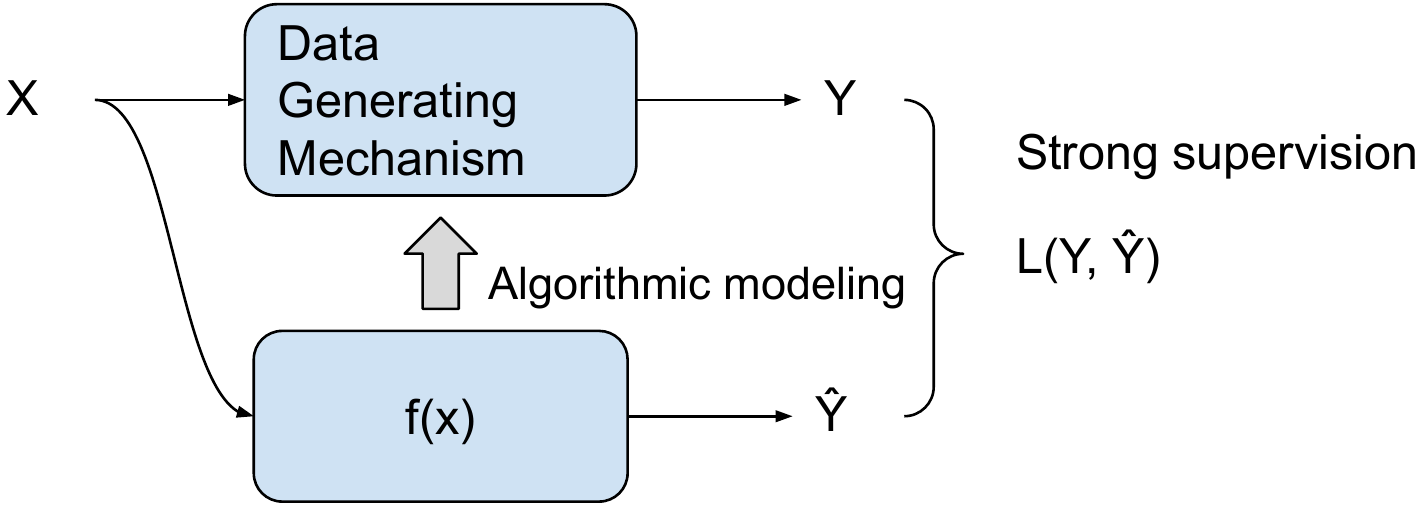}
    \caption{Strong supervision drives performance improvement in supervised learning.}
    \label{fig:strong}
\end{figure}

The past two decades have witnessed the rapid expansion and success of the algorithmic modeling culture. From self-driving cars \citep{bojarski2016end} to virtual assistants \citep{devlin2018bert}, complicated algorithmic models such as \emph{deep neural networks} (DNNs) have demonstrated their potential for leveraging today's big data and affordable high-performance computational resources in producing predictions that are comparable to human performance. However, training such algorithms to attain impressive performance relies heavily on a large volume of training data with high-quality labels (see Figure~\ref{fig:strong}), which are often expensive or even unavailable in many real-world applications. In particular, such a strong supervision becomes substantially scarcer in application domains that are more specialized, such as healthcare \citep{miotto2018deep} and ecological studies \citep{christin2019applications, tanglarge}, where domain expertise is vital in data labeling. As a result, practical challenges due to the lack of strong supervision in many real-world applications significantly limit the applicability and generalization of algorithmic models.

\emph{Weakly supervised learning} (WSL) \citep{zhou2018brief} addresses the more realistic setting when supervision is available but weak under various practical scenarios. It expands the reach of conventional supervised learning and has garnered a lot of interests in applications \citep[e.g.][]{jorgensen2008multiple, oquab2015object, peyre2017weakly}. In algorithmic modeling, strong supervision comes from a large set of accurately labelled data. Such a supervision may be weakened in approximately three ways: incomplete supervision, inexact supervision, and inaccurate supervision \citep{zhou2018brief}. 

\begin{figure}[h]
\centering
\vspace{.3cm}
\includegraphics[height=0.18\textheight]{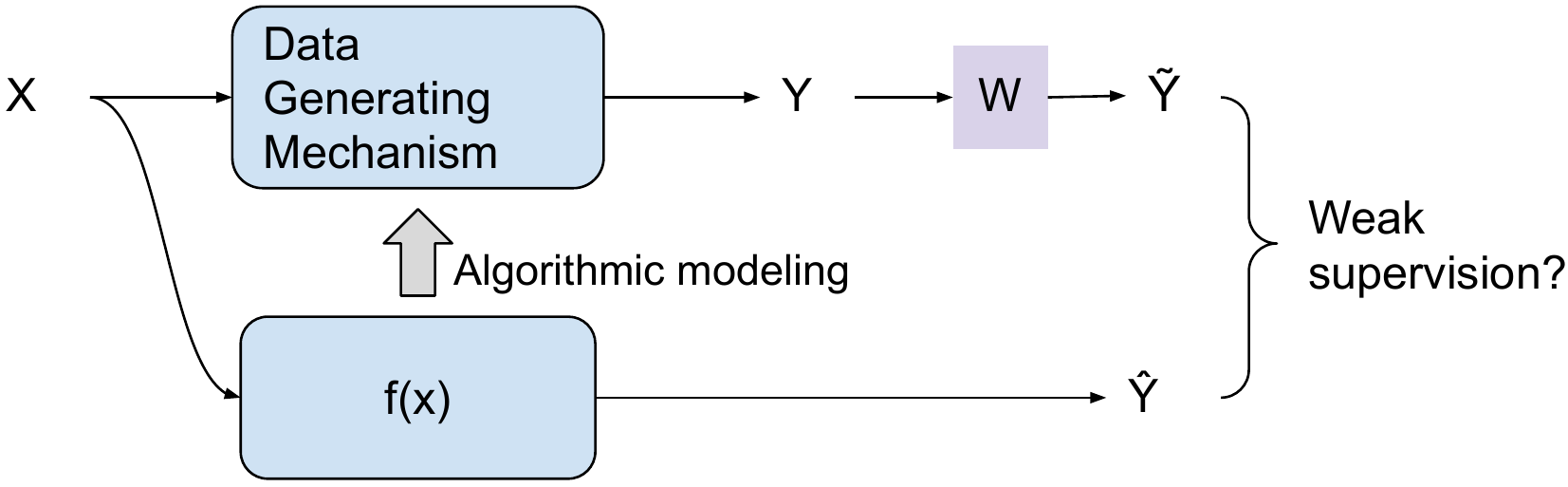}
    \caption{Weakly supervised learning poses new challenges for the algorithmic modeling framework. }
    \label{fig:weak-issue}
\end{figure}

Let $\mathbf{X}$ be the input features. Let $\mathbf{Y}$ be the outcome of interest. When $\mathbf{Y}$'s values are available in the training data as labels, they provide strong supervision for the algorithmic modeling, through a loss function $L(\mathbf{Y}, \hat{\mathbf{Y}})$ (Figure~\ref{fig:strong}). In practice, the exact and accurate values of $\mathbf{Y}$ are often unavailable in the training data. Instead, let $\widetilde{\mathbf{Y}}$ be the observed (weak) labels in the training data. Here, we introduce a unified notation, $\mathbf{W}$, for the generating mechanism of the weakened supervision $\widetilde{\mathbf{Y}}$. Using the above notation, the framework of WSL is summarized in Figure~\ref{fig:weak-issue}. WSL shares the same learning goal with methods of the algorithmic modeling culture in \citet{breiman2001statistical}, that is, train a function $f(\mathbf{X})$ such that $\mathbf{Y}$ can be accurately predicted or approximated by $f(\mathbf{X})$. Here, the challenges arise mostly from the lack of strong labels $\mathbf{Y}$ and the need to create effective supervision based on the observed $\widetilde{\mathbf{Y}}$.

Directly applying algorithmic modeling to data with weakened training labels {\em without} considering the weak supervision generating mechanism $\mathbf{W}$ could lead to results that are unstable and overfitted \citep{frenay2013classification, van2020survey}. Take semi-supervised learning as an example, which can be thought of as a special case of {\em incomplete supervision} \citep{zhou2018brief}. Most algorithms in the field of semi-supervised learning rely on the assumption that the labels are missing complete at random. When this assumption is violated in real data, semi-supervised learning algorithms may actually degrade the learning performance, compared to applying supervised learning methods directly on the labeled portion of the dataset \citep{Zhu08}. Another example is training DNNs with noisy training labels, i.e., {\em inaccurate supervision}. \citet{Zhang-et-al17} provided empirical results showing that DNNs can fit training data with randomly shuffled labels arbitrarily well. Not surprisingly, the generalization performance of the trained DNNs on test sets was no better than random guessing. Even for the relatively shallow tree ensemble models, numerical experiments have shown that the adaptive boosting algorithm (AdaBoost) would disproportionately focus on learning mislabeled instances when label noises exist \citep{dietterich2000experimental}. Therefore, an algorithmic modeling framework under weak supervision needs to explicitly acknowledge the weakening mechanism $\mathbf{W}$.

The entire promise of WSL lies within the assumption that the weak labels $\widetilde{\mathbf{Y}}$ in the training data carry partial information of $\mathbf{Y}$ through the weak supervision generating mechanism $\mathbf{W}$. Most current methods in WSL assume that the mapping from $\mathbf{Y}$ to $\tilde{\mathbf{Y}}$ by $\mathbf{W}$ is independent of the features $\mathbf{X}$ and the true labels $\mathbf{Y}$.
A more realistic scenario, however, would be that the mechanism of $\mathbf{W}$ may be dependent of both $\mathbf{X}$ and $\mathbf{Y}$. Consider the joint distribution of the observed weakened labels $\tilde{\mathbf{Y}}$ and the features $\mathbf{X}$,
\begin{equation}
P(\tilde{\mathbf{Y}}, \mathbf{X}) = \sum_Y P(\tilde{\mathbf{Y}} | \mathbf{Y}, \mathbf{X}) P(\mathbf{Y} | \mathbf{X})P(\mathbf{X}).
\label{eq:joint}
\end{equation}

\begin{figure}[h]
\centering
\includegraphics[height=0.27\textheight]{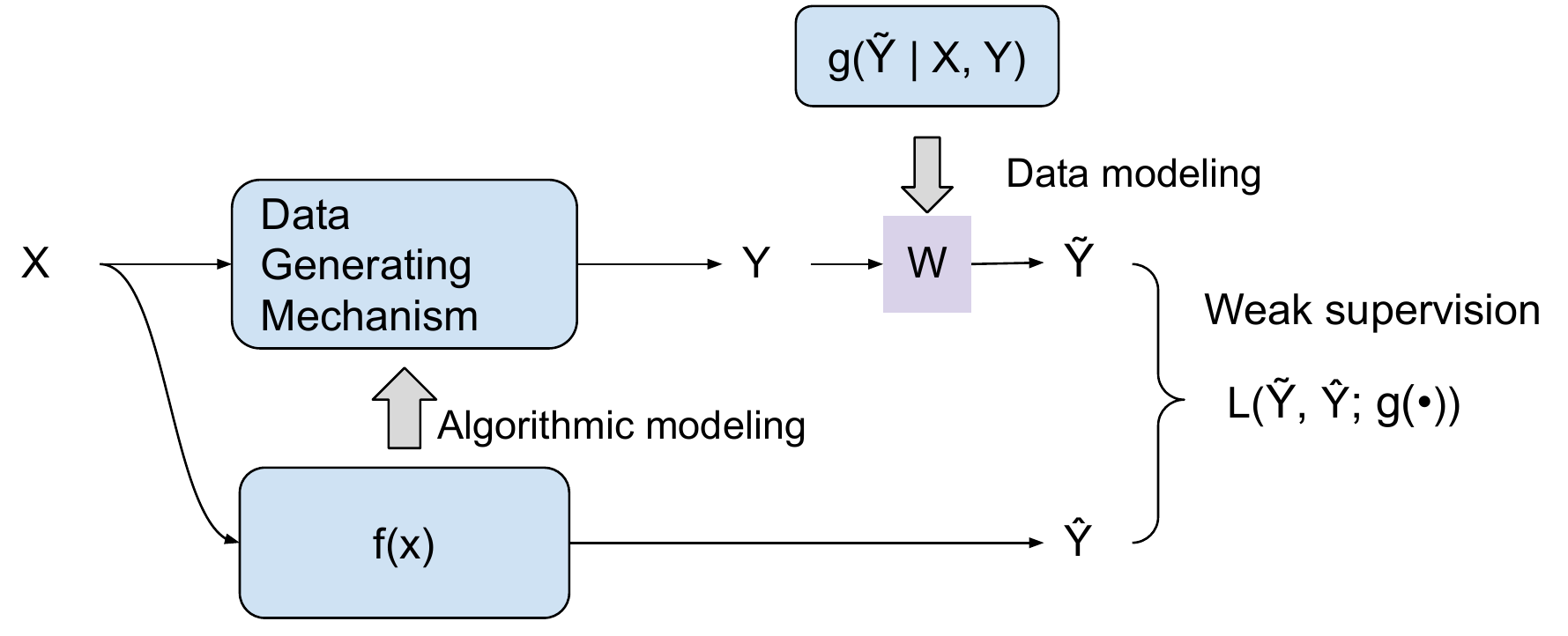}
    \caption{Weakly supervised learning requires a fusion of data modeling thinking into the algorithmic modeling framework. }
    \label{fig:weak}
\end{figure}

%
As shown in Figure~\ref{fig:weak-issue}, our learning goal remains to be fitting $P(\mathbf{Y} | \mathbf{X})$, the unknown {\em data generating mechanism}, with a model $f(\mathbf{X})$, even when we lack direct observations of $\mathbf{Y}$. To allow information in $\tilde{\mathbf{Y}}$ be passed onto the learning of $f(\mathbf{X})$, it is critical to model $\mathbf{W}$ in Figure~\ref{fig:weak-issue}, i.e., $P(\tilde{\mathbf{Y}} | \mathbf{Y}, \mathbf{X})$ on the right-hand side of Equation~(\ref{eq:joint}). In Figure~\ref{fig:weak}, we introduce an overly generalized notation $g(\tilde{\mathbf{Y}} | \mathbf{Y}, \mathbf{X})$ to encapsulate models and approaches for the mechanism $\mathbf{W}$. In practice, characterizing $P(\Tilde{\mathbf{Y}}| \mathbf{Y}, \mathbf{X})$ could be challenging as information is often scarce. In fact, without additional information beyond the training data, it is not possible to effectively leverage the weak supervision that is offered by $\tilde{\mathbf{Y}}$ \citep{frenay2013classification, zhou2018brief}. In the weakly supervised learning literature, additional information for constructing $g(\tilde{\mathbf{Y}} | \mathbf{Y}, \mathbf{X})$ has been introduced in the form of assumptions on $\mathbf{W}$ and/or small sets of data with observed $\mathbf{Y}$. This is primarily motivated by the need for transparency and interpretability for $g(\tilde{\mathbf{Y}} | \mathbf{Y}, \mathbf{X})$ to incorporate prior knowledge into ``end-to-end'' modeling frameworks. In other words, it is desirable to have the modeling of $\mathbf{W}$ be ``assumption-driven" rather than ``data-driven'' or ``accuracy-driven'', which creates a role for the {\em data modeling} culture within an {\em algorithmic modeling} framework (Figure~\ref{fig:weak}). In the weakly supervised learning literature, there has been some progress made to address each type of weak supervision. 

For {\em incomplete supervision} where labels are only available for a small subset of training data, active learning algorithms \citep{settles2009active} attempt to better extract label information by ``actively'' asking an ``oracle" (e.g., a human annotator) for queries of selected unlabeled instances. This framework has been widely used in image classification \citep{joshi2009multi, kapoor2007active, li2013adaptive}. Assuming the existence of an ``oracle'', the key component of active learning is to choose the most ``valuable'' instance to query. To this end, measures of informativeness and representativeness of individual observations have been proposed \citep{settles2009active}. For example, Bayesian active learning methods estimate the expected improvement of each instance query through nonparametric models such as Gaussian process and Monte Carlo estimations \citep[e.g.][]{gal2017deep, kapoor2007active, roy2001toward}. As another approach to incomplete supervision, semi-supervised learning algorithms \citep{chapelle2009semi, zhu2005semi} utilize the unlabeled training data as well as labeled data to improve prediction accuracy. Transductive methods were proposed to obtain label prediction for unlabelled data points \citep{van2020survey}, including the use of probabilistic models, such as Markov random fields and Gaussian random fields, for label assignments \citep[e.g.][]{shental2005semi, wu2012learning, zhu2002towards}. 


{\em Inexact supervision} addresses the situation where the given labels are at coarser scales than desired. For example, in many real-world object segmentation tasks, only image-level training labels are available, while the task is to localize each object. Multi-instance learning \citep{zhou2007multi} was such an example with a {\em bag-of-instances} setup: instances $x_{ij}$ are organized in bags $X_{j}$, and the labels in the training set are only given at the bag level. A common assumption for this task is that the bag-level class probability is the maximum of all the instance-level class probabilities within the bag. This assumption bridges the gap between instance predictions and observed bag labels. Another example is the {\em concept labeling} method \citep{chenthamarakshan2011concept}, which assumes a soft bag-instance structure. In their Bayesian modeling framework, each document (instance) $\mathbf{X}$ has a distribution $P(\mathbf{V} | \mathbf{X})$ over the concepts of the ontology $\mathbf{V}$ (bag). It is assumed the outcome variable of interest $\mathbf{Y}$, categories, is conditionally independent of document content $\mathbf{X}$, when conditioning on the oncology concept $\mathbf{V}$. Consider $P(\mathbf{Y} | \mathbf{X}) = \sum_{\mathbf{V}} P(\mathbf{Y}, \mathbf{V} | \mathbf{X}) = \sum_{\mathbf{V}} P(\mathbf{Y}|\mathbf{V}) P(\mathbf{V} | \mathbf{X})$. As a result, by separately modeling the document-to-concept distribution $P(\mathbf{V}|\mathbf{X})$ and the concept-to-class distribution $P(\mathbf{Y} | \mathbf{V})$, the instance-level document label predictions $P(\mathbf{Y}|\mathbf{X})$ can be obtained.

{\em Inaccurate supervision} concerns the situation where labels are a noisy version of the ground truth. To learn with noisy labels, many algorithms make the assumption that the noises are randomly generated. \cite{brodley1999identifying} proposes to first identify the potentially mislabeled instances and perform label correction. \citet{Northcutt-et-al21} proposed the {\em Confidence Learning} framework that iteratively determines which labels are more likely to be the contaminated ones, based on an estimated joint distribution of true label $\mathbf{Y}$ and observed label $\widetilde{\mathbf{Y}}$. The {\em data programming} approach proposed by \cite{ratner2016data} is a paradigm for integrating noisy labels from multiple sources, and deriving a better training set using a dependency graph that incorporates different assumptions on the weak supervision generating mechanisms. 



For any WSL framework, optimizing the generalization performance of the learned model $f(\mathbf{X})$, for $P(\mathbf{Y}|\mathbf{X})$, remains the main goal. However, it is important to consider the practical issues caused by the imperfection of available data and construct ``end-to-end" learning frameworks that take raw training data and deliver reliable final models. In this comment, we argue that the scarcity of strong supervision in many real-world applications calls for a fusion of modeling cultures that allow creative combinations of assumption-driven and data-driven approaches. There are many open problems and challenges that remain to be further explored. In particular, in many real-world learning tasks, all the above weak supervision scenarios may apply at the same time (e.g., noisy and inexact labels are only available on a small subset, as seen in \citet{tanglarge}). Most of the existing weakly supervised learning methods focus only on a single type of weak supervision.
%
%
As a result, the weak supervision generation mechanism $P(\widetilde{\mathbf{Y}} |\mathbf{Y}, \mathbf{X})$ is usually over-simplified in practice. Much of the statistical literature from the data modeling culture, e.g., robust statistics and methods for missing data, may find application in end-to-end workflows of weakly supervised learning. In addition, many current methods in WSL incorporate assumptions on the weak supervision in an ad hoc fashion. For the same reasons that have led to the lack of strong supervision in the training data, it is also impractical to assume that one can validate the learning framework using prediction accuracy on test data alone. Systematic model checking with respect to the weak supervision generating mechanism $\mathbf{W}$ is needed. 




\newpage

\vskip 0.2in
\bibliography{Zheng}
\end{document}